\documentclass[letterpaper]{article} 
\usepackage{aaai2026}  
\usepackage{times}  
\usepackage{helvet}  
\usepackage{courier}  
\usepackage[hyphens]{url}  
\usepackage{graphicx} 

\usepackage{amsmath}
\usepackage{natbib}  
\usepackage{booktabs}
\usepackage{multirow}
\usepackage{caption} 
\usepackage{algorithm}
\usepackage{algorithmic}
\usepackage{amssymb} 
\usepackage{newfloat}
\usepackage{listings}
\DeclareCaptionStyle{ruled}{labelfont=normalfont,labelsep=colon,strut=off} 
\floatstyle{ruled}
\newfloat{listing}{tb}{lst}{}
\floatname{listing}{Listing}
\newcommand{\sysname}{WeatherEdit}

\title{\sysname: Controllable Weather Editing with 4D Gaussian Field}

\author{
Chenghao Qian\textsuperscript{1}\quad
Wenjing Li\textsuperscript{1}\quad
Yuhu Guo\textsuperscript{2}\quad
Gustav Markkula\textsuperscript{1}\\
\textsuperscript{1}University of Leeds \quad
\textsuperscript{2}Carnegie Mellon University
}

\usepackage{bibentry}

\makeatletter
\def\showauthors@on{T}  
\makeatother

\begin{document}

\maketitle

\begin{abstract}
In this work, we present \sysname, a novel weather editing pipeline for generating realistic weather effects with controllable types and severity in 3D scenes. 
 Our approach is structured into two key components: weather background editing and weather particle construction. 
 For weather background editing, we introduce an all-in-one adapter that integrates multiple weather styles into a single pretrained diffusion model, enabling the generation of diverse weather effects in 2D image backgrounds.
 During inference, we design a Temporal-View (TV-) attention mechanism that follows a specific order to aggregate temporal and spatial information, ensuring consistent editing across multi-frame and multi-view images. 
 To construct the weather particles, we first reconstruct a 3D scene using the edited images and then introduce a dynamic 4D Gaussian field to generate snowflakes, raindrops and fog in the scene. 
The attributes and dynamics of these particles are precisely controlled through physical-based modelling and simulation, ensuring realistic weather representation and flexible severity adjustments. Finally, we integrate the 4D Gaussian field with the 3D scene to render consistent and highly realistic weather effects. Experiments on multiple driving datasets demonstrate that \sysname~can generate diverse weather effects with controllable condition severity, highlighting its potential for autonomous driving simulation in adverse weather. See project page at \textit{https://jumponthemoon.github.io/w-edit/}.
\end{abstract}


\section{Introduction}

3D weather editing (3D-WE) aims to generate multi-weather conditions in 3D scenes based on user instructions.
It is a crucial task in computer vision with applications in autonomous driving, augmented reality, and virtual scene synthesis \cite{Almalioglu2022}. 
While existing methods range from 2D-based image techniques \cite{Valanarasu_2022_CVPR, Zhang_2023_CVPR} to 3D neural rendering \cite{Li2023ClimateNeRF}, developing a unified framework that enables multiple realistic weather effects and controllable severity remains a significant challenge.
For instance, style transfer methods can modify weather backgrounds but are typically limited to a single effect. Diffusion models~\cite{10656364} enable multi-weather synthesis; however, they often lack fine-grained control and struggle with spatial and temporal consistency across multiple images~\cite{9565008}. More recently, ClimateNeRF~\cite{Li2023ClimateNeRF} utilized Neural Radiance Fields (NeRF) to model snow accumulation and flooding in 3D space. However, this method is limited to static weather events and leaves the overall weather tone unchanged, lacking full-scene atmospheric realism.

\begin{figure}
  \centering
  \includegraphics[width=\linewidth]{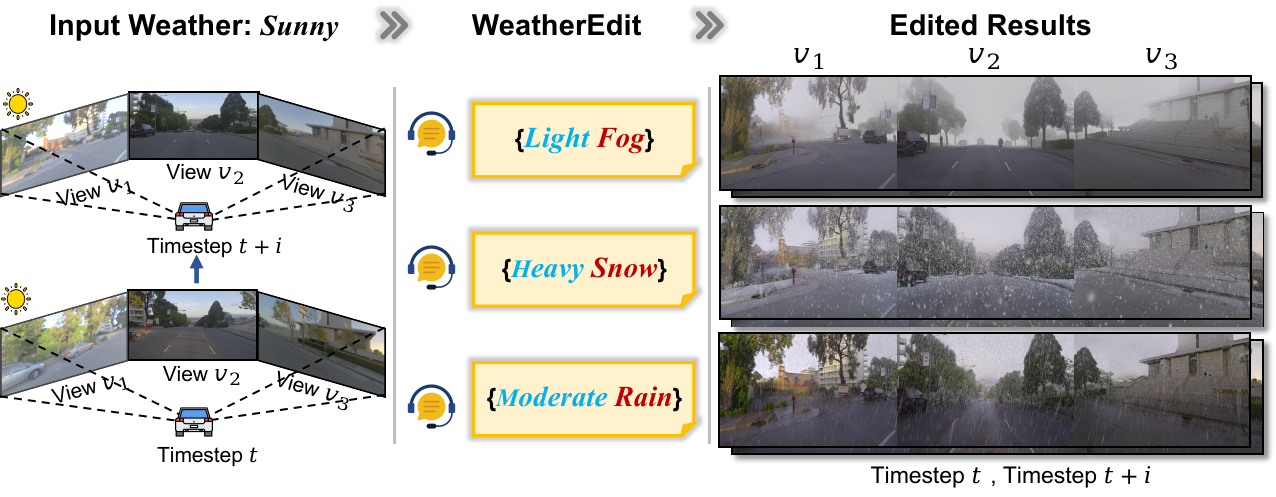}
  \captionof{figure}{\textbf{Realistic, Controllable Weather Editing.} 
Given input sunny weather images from multi-view cameras across multiple timesteps, \sysname~ can generate high-fidelity scenarios under diverse weather conditions (\textit{e.g.}, foggy, snowy, and rainy) with various severity levels (\textit{e.g.}, light, moderate, heavy) based on user instructions.}
  \label{fig:Teaser}
  \vspace{-6mm}
\end{figure}
In 3D-WE, achieving both \textbf{realism} and \textbf{controllability} is important. Realism ensures high-quality weather effects and we consider it involves three aspects:
(R1) \textit{Distortion-free editing}: The editing results should be free from visual distortions and unnatural artifacts.
(R2) \textit{Temporal-view consistency}: Weather effects and scene elements should maintain consistency across both time steps and different viewpoints.
(R3) \textit{Dynamic weather effects}: Weather particles, such as snowflakes and raindrops, should be realistically simulated to reflect natural weather dynamics.
On the other hand, controllability ensures flexible and fine-grained editing of weather effects, enabling users to customize various aspects of the scene, including:
(C1) \textit{Type}: The system should support the selection and transition between different weather types (\textit{e.g.}, rainy, snowy, foggy).
(C2) \textit{Event}: The system should allow users to specify whether the weather effect is static (e.g., previously fallen snow) or ongoing (e.g., falling snow).
(C3) \textit{Severity}: Users should be able to adjust the intensity of weather conditions, such as changing light rain to heavy rain or increasing fog density.
Unfortunately, current approaches have not explicitly considered the above aspects, leading to limitations in realism and controllability.

In this paper, we consider the above points and introduce a novel weather editing framework named WeatherEdit that enables realistic, and controllable weather generation in 3D scenes.
WeatherEdit follows a progressive 2D-to-4D transformation process, starting with 2D image-based background editing, transitioning to 3D scene reconstruction, and culminating in 4D
dynamic weather simulation.
This is achieved through a two-step process: \textbf{weather background editing} and \textbf{weather particle modeling}. 

In weather background editing, we first introduce an efficient all-in-one adapter that integrates multiple weather styles into a single model, allowing for the fine-tuning of a diffusion model to achieve distortion-free rendering (R1) and flexible weather type control (C1). 
%
%
Second, to ensure temporal-view consistency (R2), we introduce a parameter-free attention mechanism that aligns different viewpoints simultaneously and the same viewpoint over time.
For weather particle modeling, we propose a novel 4D Gaussian field, representing weather particles as 3D Gaussians with attributes such as color and density while simulating their motion. This enables dynamic weather effects (R3) and severity (C3). Rather than simulating an infinite field for full-scene coverage, we optimize efficiency by recycling particles within a compact field and aligning it with the rendering camera, maintaining both realism and performance.

The final weather effect is rendered from the 4D Gaussian field combined with the 3D scene reconstructed using edited weather background images. This flexible combination allows control over weather events, such as a wet road without rainfall (static) or a heavy snowstorm with limited visibility (dynamic), enabling adjustable event (C2).

%
%

\begin{figure*}[h!]
    \centering
    \includegraphics[width=.96\linewidth]{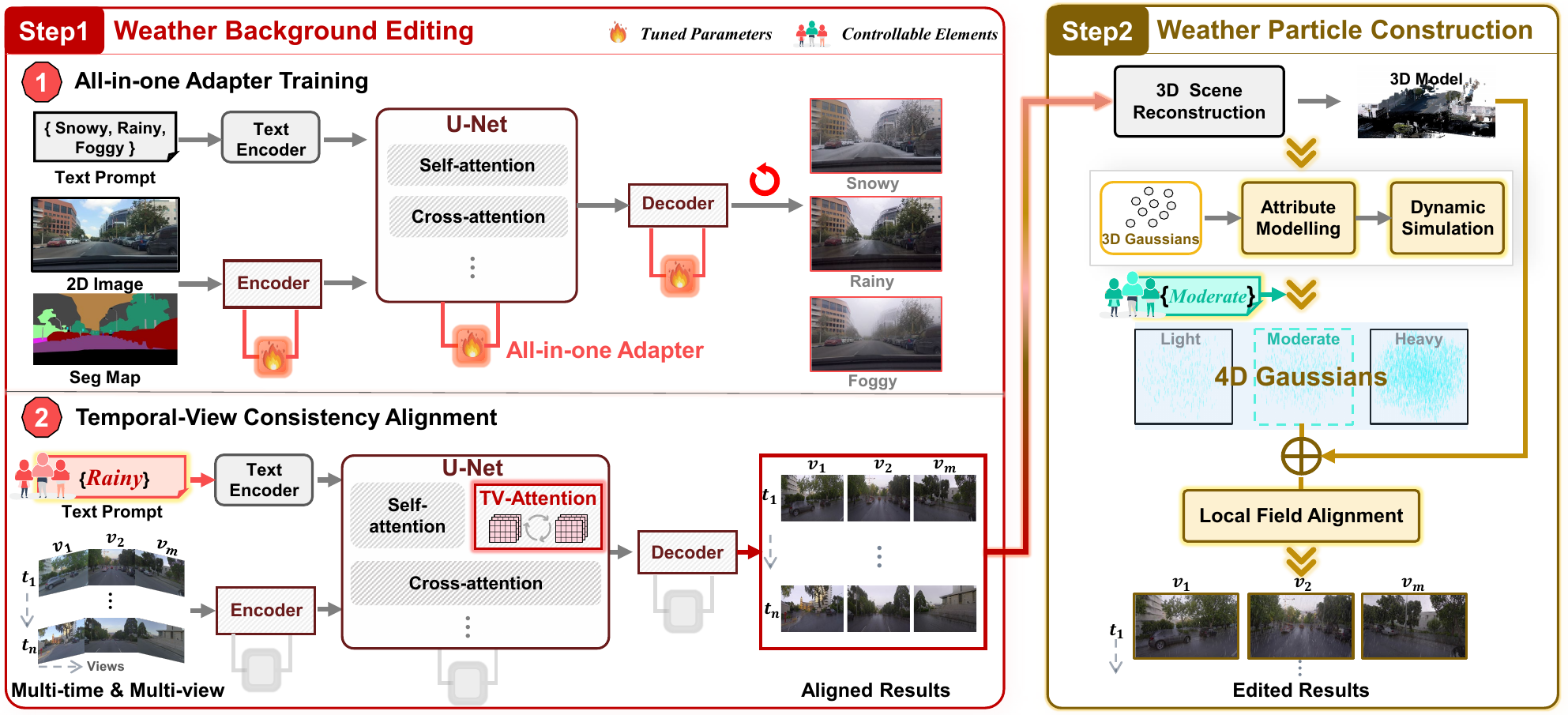}
    \caption{\textbf{Overview of the WeatherEdit Framework.} (a) Weather Background Editing: An all-in-one adapter enables the synthesis of diverse weather effects (e.g., snow, rain, fog) from text prompts and segmentation maps. Inference takes multi-frame and multi-view images as input, using temporal-view (TV) attention to ensure consistency. (b) Weather Particle Construction: A 4D Gaussian field, initialized from 3D Gaussians, undergoes attribute modeling and dynamic simulation, allowing user-controlled weather intensity (e.g., light, moderate, heavy). Local field alignment ensures seamless integration of the reconstructed 3D scene with dynamic weather effects.
}
    \label{fig:Method_overview}
\end{figure*}

The main points of this paper are summarized below.

\begin{enumerate}
    \item Based on our analysis of weather editing characteristics, we introduce WeatherEdit, a comprehensive and efficient framework for realistic and controllable weather generation.
    Compared with existing methods which focus on either background editing or static weather effects, a progressive 2D-to-4D transformation process in WeatherEdit  enhances adaptability across a wider range of scenarios.

    \item  We introduce an all-in-one adapter 
    to enable a diffusion model for multi-weather (snowy, rainy and fog) synthesis and a Temporal-View attention to ensure consistent editing across multi-frame and multi-view.
    
    \item We design a 4D Gaussian field for weather particle modeling, enabling plausible simulation of raindrops, snowflakes, and fog with controllable severity.
    
    \item We demonstrate \sysname’s effectiveness in generating realistic, consistent, and controllable weather effects in 3D driving scenes (shown in Figure~\ref{fig:Teaser}), showcasing its applicability to real-world scenarios.
\end{enumerate}

\section{Related Work}

\subsection*{Weather Effect Synthesis}Existing methods can be categorized into 2D-based and 3D-based approaches. 2D-based methods utilize generative models \cite{zhu2017unpaired,parmar2024one} to synthesize weather effects in single images. \cite{hwang2022weathergan, schmidt2019visualizing} use CycleGAN \cite{zhu2017unpaired} to edit weather conditions, while 
\cite{schmidtclimategan} applies GAN-based inpainting to generate flooding with water reflections. \cite{hahner2019semantic} uses a graphical model to produce fog effects. Conventional 3D-based methods \cite{stomakhin2013material,zsolnai2022flow,feldman2002modeling, haas2014history} use graphical simulation to model weather in 3D space. More recently, \cite{Li2023ClimateNeRF} has introduced neural rendering techniques for creating weather backgrounds but it is limited to post-weather event.


\subsection*{Diffusion Model for Image Editing} Diffusion models \cite{ho2020denoising,meng2021sdedit,rombach2021highresolution} have significantly advanced image editing, enabling flexible and controllable modifications. Methods such as ControlNet \cite{Zhang_2023_ICCV} and T2I-Adapter \cite{10.1609/aaai.v38i5.28226} introduce trainable modules to guide the generation process, allowing for fine-grained control over diffusion-based outputs. InstructPix2Pix \cite{10204579} enables interactive editing with textual instructions by trained on image-text pairs. TurboEdit \cite{deutch2024turboedittextbasedimageediting} employs a pseudo-guidance approach to enhance edit magnitude without introducing artifacts.

Although these methods support diverse styles, they are not specifically designed for weather editing, often distorting content or generating overly artistic effects. Moreover, they primarily focus on single-image editing, lacking spatial and temporal consistency across multiple frames and views. These limitations restrict their ability to generate realistic and consistent weather effects within the same scene.
\subsection*{3D Scene Reconstruction. } Neural Radiance Fields (NeRF) \cite{10.1007/978-3-030-58452-8_24}and 3D Gaussian Splatting (3DGS) \cite{kerbl3Dgaussians} are leading techniques for 3D scene reconstruction from multi-view images. NeRF encodes scenes implicitly with a neural network, while 3DGS represents them explicitly as 3D Gaussians, enabling faster rendering. Their versatility allows applications in dynamic scene reconstruction \cite{chen2025omnire,yan2023nerf}, content creation \cite{liu2024humangaussian,lin2023magic3d}, and artifact removal \cite{qian2024weathergs}. While \cite{Li2023ClimateNeRF} extends NeRF to model smog, snow, and floods for weather simulation but focuses only on static weather conditions, leaving dynamic weather phenomena unaddressed.

\section{Methods}
In order to generate realistic and controllable multi-weather scenes in 3D, we propose a novel framework, WeatherEdit, which follows a progressive 2D-to-4D transformation process. As illustrated in Figure~\ref{fig:Method_overview}, \sysname~ adopts a two-step weather editing pipeline: weather background editing and weather particle construction. The first step generates distortion-free, spatially and temporally aligned 2D images with edited backgrounds, while the second step dynamically models weather particles with various types and severity levels. The joint association of these two steps enables the flexible generation of multi-weather scenes.

\subsection{Weather Background Editing}
\label{sec:2D_edit}
To support multiple editing styles, we develop a unified adapter to finetune the diffusion model and further enhance the results through semantically conditioned inputs. Additionally, we introduce a novel Temporal-View Attention mechanism to ensure consistency in the edited results. 

\begin{figure}[t!]
    \centering
    \includegraphics[width=\linewidth]{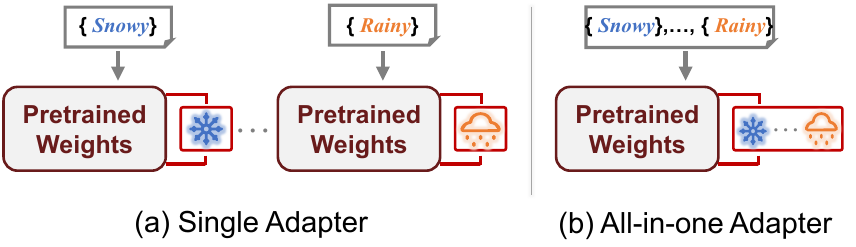 }
    \caption{\textbf{Single-style Adapters vs. All-in-One Adapter.} (a) Utilizes separate adapters for each weather style (e.g., snow, rain, fog), while (b) employs a unified all-in-one adapter for efficient multi-style adaptation. Blue box highlights adapters to be trained.}
    \label{fig:Method_lora}
\end{figure}

\noindent \textbf{All-in-one Adapter.}
Existing diffusion-based methods often suffer from scene deformations and excessive stylization, making them unsuitable for precise and realistic weather editing. Although \cite{parmar2024one} achieves effective style editing by fine-tuning the models with multiple single-style LoRA \cite{hu2022lora} adapters (shown in Figure~\ref{fig:Method_lora} (a)), it requires training a separate model for each style, resulting in increased computational cost and limited flexibility.

To address this, we develop an all-in-one LoRA adapter that injects multiple weather styles into a single model (as shown in Figure~\ref{fig:Method_lora} (b)), eliminating the need for training multiple ones. 
Let \( W_0 \in \mathbb{R}^{d \times k} \) be the pre-trained weights, \( x \) the input image, and \( i \) the weather style (\textit{e.g.}, snowy, rainy, foggy) specified by text prompt. 
Each style is associated with a low-rank matrix \( L_i \). The adapted model’s forward pass is then computed as \( h = W_0 x + L_i x \). 


Moreover, since weather effects depend on scene semantics (e.g., snow on trees, rain on roads), we enhance editing by conditioning the model on semantic segmentation maps. Given an input image \( x \) with its corresponding segmentation map \( M \), we encode the latent representation as \( \mathbf{z}_0 = \mathcal{E}(x, M) \). The model fine-tuning objective is then formulated as:
\begin{equation}
    \mathcal{L}(\Delta \theta) = \mathbb{E}_{\epsilon, t} \left[ \| \epsilon - \epsilon_{\theta + \Delta \theta} (\sqrt{a_t} \mathbf{z}_0 + \sqrt{1 - a_t} \epsilon, t, p) \|^2 \right],
\end{equation}

\noindent where \( \epsilon \) is the sampled Gaussian noise, \( \epsilon_{\theta + \Delta \theta} \) is the UNet model with LoRA, and \( p \) is the text prompt. This formulation ensures editing results are semantically relevant, enabling precise and context-aware weather editing.

\noindent \textbf{Temporal-View Consistency Alignment.}
\begin{figure}[t!]
    \centering
    \includegraphics[width=.9\linewidth]{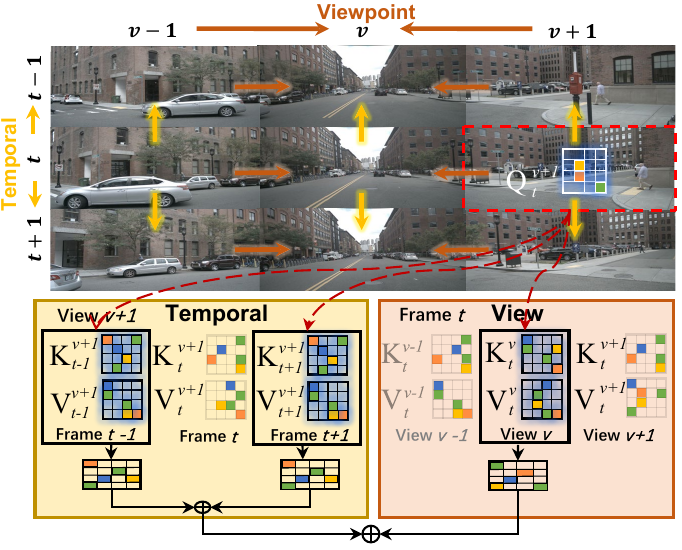}
    \caption{\textbf{Temporal-View  Attention for Consistent Weather Editing.} The TV-attention aggregates information across both adjacent frames and center viewpoints to ensure multi-frame and multi-view consistency. The top section illustrates frame and viewpoint relationships, indicated by directional arrows, while the bottom section visualizes the attention operations.}
    \label{fig:TV-Attention}
    \vspace{-3mm}
\end{figure}
After fine-tuning the model, 
 multi-time and multi-view images are processed to achieve background scene editing. 
However, processing each image independently can lead to inconsistencies across time steps and viewpoints due to the inherent stochasticity of generative models. 
To address this, we propose a Temporal-View (TV-) attention mechanism (as shown in Figure~\ref{fig:TV-Attention}) to enable our model to integrate information from adjacent frames and key viewpoints. 


In multi-view driving datasets, cameras are typically positioned in a structured left-front-right arrangement. 
The front camera captures overlapping content with both the left and right cameras, providing shared information across views. 
To leverage this, we design the view attention to allow the left and right cameras to query the center (front) view:

\begin{equation}
    \text{S\_Attn}^v_t = \text{Softmax} \left( \frac{ Q^v_t \cdot K^{v,T}_t }{\sqrt{d}} \right) V^{v}_t ,
\end{equation}

\noindent where \( Q^v_t \) is the query from view \( v \), and \( K^v_t, V^v_t \) are the key and value derived from the center view \( v \). This ensures that each side viewpoint aggregates information from the center view, maintaining result consistency across different perspectives.

To preserve temporal coherence, each frame \( t \) queries its adjacent frames (\( t-1 \) and \( t+1 \)), allowing it to integrate contextual information from both past and future frames. The temporal attention  is defined as:

\begin{equation}
    \text{T\_Attn}^v_t = \text{Softmax} \left( \frac{ Q^v_t \cdot K^{\mathcal{T}(t),T}_v }{\sqrt{d}} \right) V^{\mathcal{T}(t)}_v ,
\end{equation}

\noindent where \( \mathcal{T}(t) = \{t-1, t+1\} \), ensuring that each frame incorporates temporal context, preventing inconsistencies across consecutive frames. The final attention is computed as a weighted sum of self-, view-, and temporal-attention, with $\lambda$ controlling the balance to ensure global coherence across frames and viewpoints:
\begin{equation}
    \text{Attn}^v_t = \lambda \cdot \text{SelfAttn}^v_t + (1 - \lambda) \cdot (\text{S\_Attn}^v_t + \text{T\_Attn}^v_t).
\end{equation}

In this way, we can effectively reduce inconsistencies caused by independent image processing, producing a coherent editing effect across both view and temporal domains.

\subsection{Weather Particle Construction}
\label{sec:3D_edit}
\begin{figure}[t!]
    \centering
    \includegraphics[width=.9\linewidth]{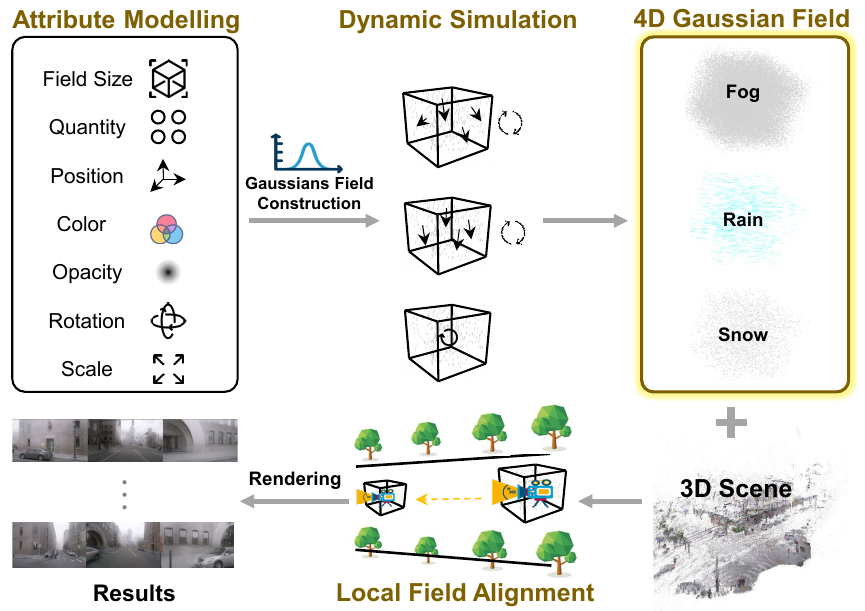}
    \caption{\textbf{Weather Particle Construction Process.} The attribute modeling stage defines weather properties to construct a 4D Gaussian field, which then undergoes dynamic simulation to model fog, rain, and snow. Local field alignment ensures seamless integration with the 3D scene, enabling realistic weather rendering. }
    \label{fig:Particle_Model}
\end{figure}

Real-world weather is inherently regional,
and we incorporate this into weather particle construction. 
Specifically, as shown in Figure~\ref{fig:Particle_Model}, we propose a \textit{4D Gaussian field}, initialized with 3D Gaussians carrying physical attributes (\textit{e.g.}, color, quantity) to model raindrops, fog, and snowflakes through attribute modeling. 
To enhance realism, we apply dynamic simulation to create realistic weather effects.

To generate a consistent falling particle effect, a large number of particles must be simulated, which is computationally expensive. To address this, we model particle behavior within a localized field, dynamically recycling particles as they exit while aligning the field with the rendering camera. This ensures a continuous and realistic weather effect while optimizing computational efficiency.

\noindent \textbf{Attribute Modeling.} Different weather particles exhibit unique characteristics. 
For instance, snowflakes are typically white with irregular shapes, while raindrops are transparent and elongated. 
To accurately capture these differences, we define a set of characteristic attributes \( A \) to distinguish different types of weather particles: $A_i = \{C_i, P_i, R_i, S_i, O_i\}$, where, \( C_i \) defines color, \( P_i \) represents position, \( R_i \) and \( S_i \) correspond to rotation and scale, and \( O_i \) signifies opacity. 
The index \( i \) refers to a specific type of weather, such as snow, rain, or fog. Additionally, we model these parameters using Gaussian distributions, expressed as \( \mathbb{G}(A_i) \), to mimic the natural randomness of real-world weather.
To compose the 4D Gaussian field, we use \( B_i \) to define the particle field size and combine it with quantity \( q \), which controls the number of weather particles to regulate density. 
The final attributes of the 4D Gaussian field, are denoted as: $W_i = \{ \mathbb{G}(A_i), q, B_i \}$.

\noindent \textbf{Dynamic Simluation.} Real-world weather is dynamic, with particles falling under gravity and wind. Instead of simulating these forces in detail, we approximate motion using a constant directional velocity, updating each particle's position as follows:

\begin{equation}
\mathbf{\textit{P}}_j(t + \Delta t) = \mathbf{\textit{P}}_j(t) + \mathbf{\textit{D}}_j \cdot \Delta t ,
\end{equation}
\noindent where 
\begin{equation}
    \mathbf{\textit{P}}_j(t) = \begin{bmatrix} x(t), y(t), z(t) \end{bmatrix}^\top,
    \mathbf{\textit{D}}_j = \begin{bmatrix} D_x, D_y, D_z \end{bmatrix}^\top
\end{equation}
represent the position and velocity of the $j$-th particle, respectively. Here, \( x(t), y(t), z(t) \) denote the spatial coordinates of the particle at time \( t \), while \( D_x, D_y, D_z \) are its velocity components along the \( x \), \( y \), and \( z \) axes.


To efficiently simulate a consistent falling particle effect, we recycle particles when they exit the predefined simulation field by resetting their positions. The position of the \( i \)-th particle, \( \mathbf{\textit{P}}_i(t) \), is updated as follows:

\begin{equation}
   \mathbf{\textit{P}}_i(t) = 
\begin{cases} 
p_\text{max} - \delta_p, & \text{if } p_i(t) < p_\text{min}, \\ 
p_\text{min} + \delta_p, & \text{if } p_i(t) > p_\text{max}, \\ 
\mathbf{\textit{P}}_i(t), & \text{otherwise}.
\end{cases} 
\end{equation}

Here, \( p_\text{min} \) and \( p_\text{max} \) denotes the field's lower and upper bounds, while \( \delta_p \) represents the offset to the field bounds.


\noindent \textbf{Local Field Alignment.} After obtaining the reconstructed 3D scene and simulated 4D Gaussian field, we combine them to render the final weather scene. In large-scale environments, the weather field must scale appropriately to ensure a consistent weather effect. 
However, as scene size increases, more particles are required, leading to higher computational costs. 
To mitigate this, we constrain the size of the 4D Gaussian field and limit the number of particles, aligning its motion with the rendering camera. 
This keeps the field relatively stationary to the camera while preserving internal dynamics and intended weather effects. 
Given camera poses \( \mathbf{T}_0 \) at \( t_0 \) and \( \mathbf{T}_t \) at \( t \), the relative transformation is:

\begin{equation}
    \Delta \mathbf{T} = \mathbf{T}_t \mathbf{T}_0^{-1}.
\end{equation}

The particle’s position, initially \( P_j(t_0) \), updates as:

\begin{equation}
    P_j(t) = \Delta \mathbf{R} P_j(t_0) + \Delta \mathbf{t},
\end{equation}

\noindent where \( \Delta \mathbf{R} \) and \( \Delta \mathbf{t} \) are the rotation and translation from \( \Delta \mathbf{T} \). In this way, we can ensure consistent weather dynamics in large-scale scenes with minimal computational cost.

\section{Experiments}
\begin{figure*}
    \centering
    \includegraphics[width=\linewidth]{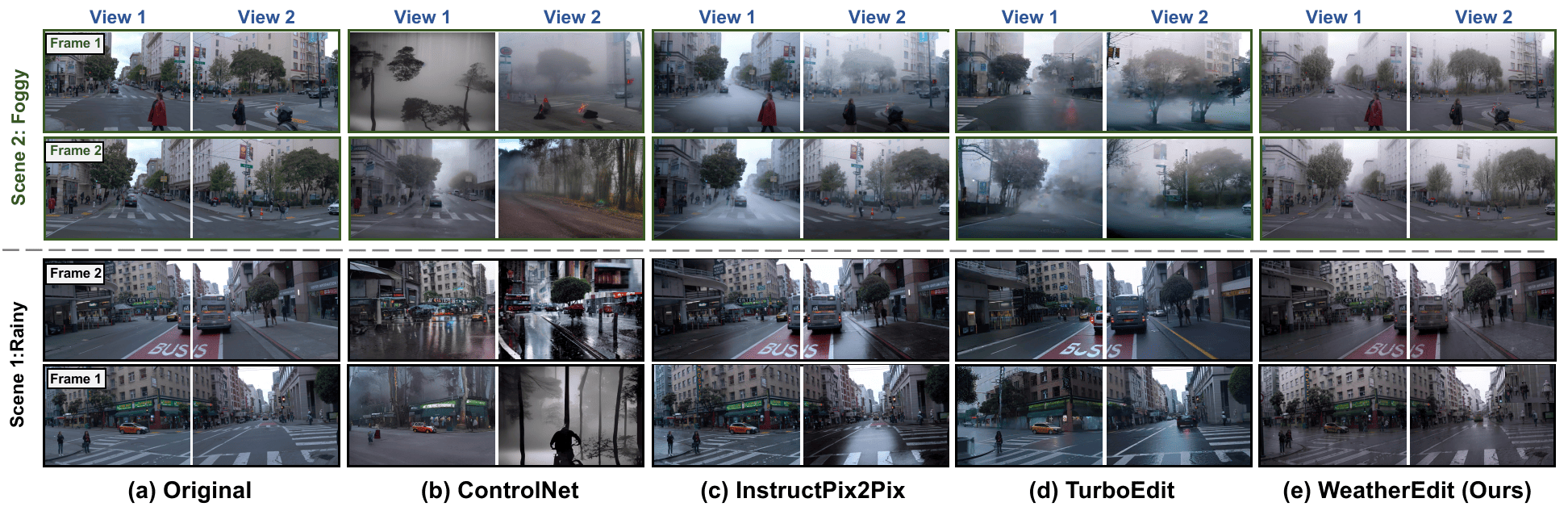}
    \caption{\textbf{Comparison with 2D-based Editing Methods.} We show (a) original images with edited results from (b) ControlNet, (c) InstructPix2Pix, (d) TurboEdit, and (e) Ours. Existing methods often suffer from over-stylization, content removal, and inconsistencies in spatial and temporal coherence. In contrast, our method preserves scene integrity while ensuring multi-view and multi-frame consistency, producing realistic and temporally coherent weather effects across viewpoints and frames.}
    \label{fig:Exp_2D}
\end{figure*}
\begin{figure*}[ht]
    \centering
    \includegraphics[width=\linewidth]{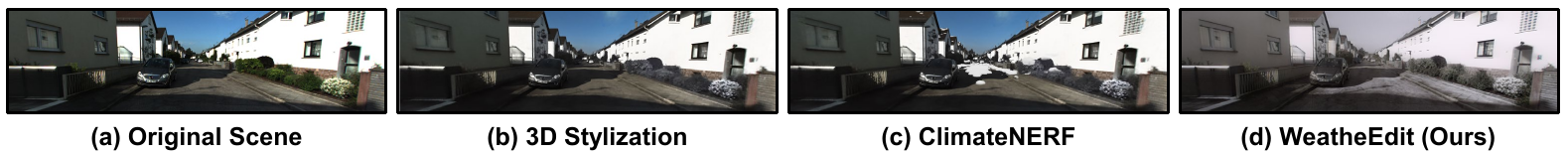}
    \caption{\textbf{Comparison with 3D-based weather synthesis methods in editing \textit{snowy}.} 3D Stylization (b) produces only a subtle snowy effect on the sidewalk, while (c) ClimateNeRF present unnatural snow cover. In contrast, our method (d) synthesizes a realistic snowy effect while ensuring a perceptually convincing atmosphere.}
    \label{fig:3D-Comparison}
    \vspace{-6mm}
\end{figure*}

\subsection{Experimental Setup}

\noindent \textbf{Dataset.} To develop the weather dataset for all-in-one adapter fine-tuning, we extract 1,237 image pairs from the BDD100K \cite{yu2020bdd100k}, MUSE \cite{brodermann2024muses}, and ACDC \cite{sakaridis2021acdc} datasets.
Each pair contains a normal-weather image and an adverse-weather image, including snowy, rainy, and foggy conditions, along with a corresponding ground-truth semantic segmentation map.
For weather editing, we selected eight scenes with multi-view image sequences from the Pandaset \cite{xiao2021pandaset}, Waymo Open Dataset \cite{Sun_2020_CVPR}, nuScenes \cite{caesar2020nuscenes}, and KITTI-360 \cite{liao2022kitti} datasets.

\noindent \textbf{Implementation details.} We implement the 2D editing model using CycleGAN-Turbo\cite{parmar2024one}, using a single RTX A6000, which takes 12 hours training and runs at 3.2 seconds per frame. The 3D scene reconstruction and 4D Gaussian field are based on OmniRe\cite{chen2025omnire} and 3DGS\cite{kerbl3Dgaussians}, with both training(\textasciitilde1 hour) and rendering(\textasciitilde0.21s) performed on a single RTX 3090.

\noindent \textbf{Evaluation Metrics.} To evaluate edited weather background quality,  we use the cosine similarity of CLIP \cite{radford2021learning} image embeddings (CLIP-S) to measure content preservation and the directional CLIP similarity \cite{gal2022stylegan} (CLIP-DS) to assess alignment with text instructions. To assess consistency across temporal and spatial dimensions, we calculate the warp error to measure temporal coherence between adjacent frames and employ the Bhattacharyya distance \cite{bhattacharyya1943measure} to quantify color distribution similarity across viewpoints.

\section{Results}
\begin{figure}[ht]
        \centering
        \includegraphics[width=\linewidth]{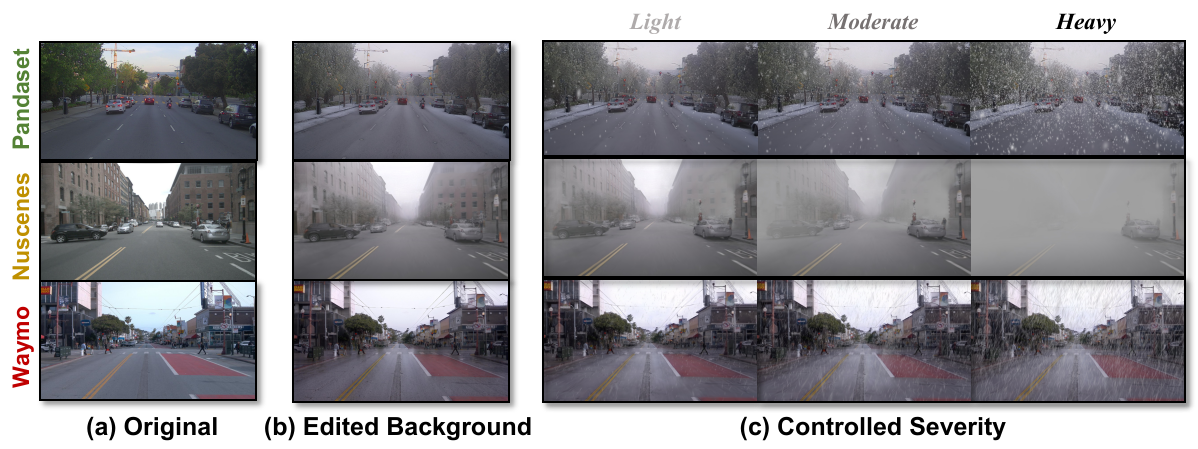}
        \caption{\textbf{Controllable Weather Effect and Severity in Driving Scenes.} WeatherEdit can modify (a) original driving scenes from Pandaset, Nuscenes, and Waymo datasets by generating snowy, foggy, and rainy effects while preserving scene structure in (b). Additionally, it enables (c) controllable weather severity, adjusting precipitation intensity from light to heavy for realistic and flexible weather simulation.}
        \label{fig:control}
        \vspace{-4mm}
\end{figure}

\subsection{Experimental Results}
We present both qualitative and quantitative comparisons of our weather background editing method against 2D-based approaches, including ControlNet \cite{Zhang_2023_ICCV}, InstructPix2Pix \cite{10204579}, and TurboEdit \cite{deutch2024turboedittextbasedimageediting}, as well as 3D-based methods, such as 3D stylization \cite{li2018closed} and ClimateNeRF \cite{Li2023ClimateNeRF}. For dynamic weather effects, we apply qualitative analysis, \textit{i.e.,} controllable weather simulation, to demonstrate the results.

\noindent \textbf{Weather Background Editing Comparison.} 
For the comparison of 2D-based methods, as shown in Figure~\ref{fig:Exp_2D}, we observe that ControlNet tends to over-stylize the image, altering its original content. InstructPix2Pix produces a reasonable effect; however, the effect is inconsistent across adjacent frames and viewpoints.
While TurboEdit accurately stylizes the image, it significantly modifies the scene content (e.g., turning pedestrians into water reflections and trees in a foggy scene). In contrast, our method achieves a realistic and coherent weather transformation while preserving scene integrity, as also indicated by the highest CLIP-DS and CLIP-S scores shown in Table~\ref{tab:Exp_2D}.


\begin{table}[t]
\setlength{\tabcolsep}{4.5pt}
\centering
\small
\begin{tabular}{lcccc}
\toprule
& \textbf{ControlNet} & \textbf{IP2P} & \textbf{TurboEdit} & \textbf{Ours} \\
\midrule
CLIP-DS Score~($\uparrow$) & 0.291 & 0.295 & 0.257 & \textbf{0.302} \\
CLIP-S Score~($\uparrow$)  & 0.594 & 0.691 & 0.676 & \textbf{0.750} \\
\bottomrule
\end{tabular}
\caption{Comparison with 2D-based weather editing.}
\label{tab:Exp_2D}
\end{table}

\begin{table}[t]
\setlength{\tabcolsep}{4.5pt}
\centering
\small
\begin{tabular}{lccc}
\toprule
& \textbf{3D Stylization} & \textbf{ClimateNeRF} & \textbf{Ours} \\
\midrule
CLIP-DS Score~($\uparrow$) & 0.225 & 0.267 & \textbf{0.283} \\
CLIP-S Score~($\uparrow$)  & \textbf{0.842} & 0.702 & 0.751 \\
\bottomrule
\end{tabular}
\caption{Comparison with 3D-based weather editing.}
\label{tab:Exp_3D}
\end{table}

For 3D-based methods, as shown in Figure~\ref{fig:3D-Comparison}, 3D stylization produces only a barely perceptible snowy effect on the vegetation, aligning with the highest CLIP-S score (shown in Table~\ref{tab:Exp_3D}), which indicates minimal change in the scene. ClimateNeRF fails to modify the overall scene tone to accurately reflect the intended weather conditions. In contrast, our approach not only synthesizes a realistic snowy effect but also adjusts the scene tone to create a coherent and perceptually convincing snowy atmosphere, as also indicated by the highest CLIP-DS score in Table~\ref{tab:Exp_3D}.

\noindent \textbf{Controllable Weather Simulation.}
A key advantage of our method over existing approaches is its ability to generate dynamic weather effects while precisely controlling both the type and severity of weather conditions. 
Additionally, we can flexibly choose whether to introduce weather particles and, if included, adjust their intensity with fine granularity. 
%
%
This allows for a range of scenarios, from a static wet road surface without active rainfall to a heavy snowstorm with dense snowfall obscuring visibility. 
%
To visualize the results, we render edited weather backgrounds and modulate weather severity. As shown in Figure~\ref{fig:control}b, backgrounds are reconstructed from 2D editing outputs, with weather type determined by the selected adapters. We then apply the 4D Gaussian field to introduce particles (Figure~\ref{fig:control}c). By adjusting particle attributes such as color, opacity, shape, rotation, and velocity, we simulate snow, rain, and fog from light to heavy. This enables flexible, physically consistent scene manipulation for diverse weather conditions.

\begin{table}[t]
\centering
\small
\begin{tabular}{lcccc}
\toprule
 & \multicolumn{2}{c}{\textbf{ACDC}} & \multicolumn{2}{c}{\textbf{MUSE}} \\
                        & w/o & w     & w/o & w \\
\midrule
\textbf{HRDA}                   & 46.3    & 58.7 (+12.4)       & 42.2    & 57.1 (+14.9)    \\
\textbf{MIC}                    & 50.0    & 59.0 (+9.0 )        & 48.8    & 55.4 (+6.8)    \\
\bottomrule
\end{tabular}
\caption{Comparison of segmentation results with and without our simulated weather data.}
\label{tab:acdc_muse_comparison}
\vspace{-4mm}
\end{table}
\noindent \textbf{Benefits for Downstream Tasks.} We validate the benefits of our simulation as data augmentation for a downstream semantic segmentation task on adverse weather dataset such as ACDC\cite{sakaridis2021acdc}, MUSE\cite{brodermann2024muses}. We simulate 1273 frames of edited weather scenes and use HRDA\cite{hoyer2022hrda} and MIC\cite{hoyer2023mic} for training. As shown in Table~\ref{tab:acdc_muse_comparison}, our simulated data boosts the performance across both datasets and models, with mIoU gains up 14.9\%. This demonstrates that our simulated scenes effectively enhance model robustness under adverse weather conditions, validating the utility of WeatherEdit for downstream tasks.

\section{Ablation Studies}
We conduct ablation studies on the semantic condition input for finetuning, temporal-view attention for image editing and modular design in weather particle construction. 

\begin{figure}[t!]
    \centering
    \includegraphics[width=0.95\linewidth]{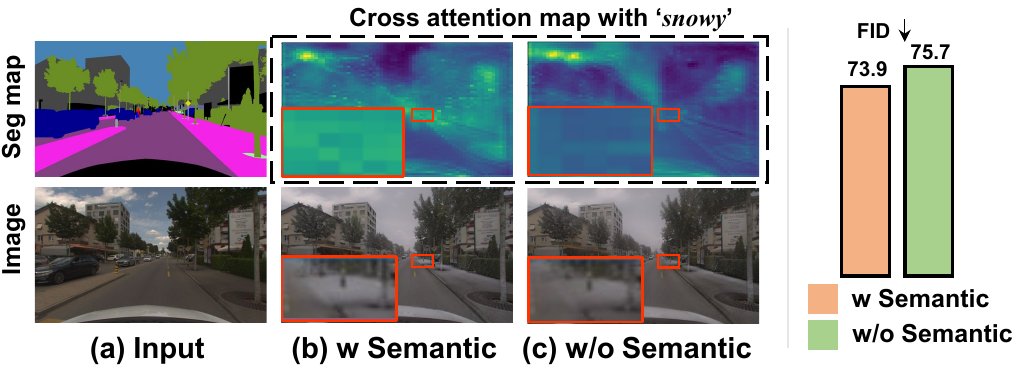}
    \caption{\textbf{Ablation on semantic conditioned input. } 
    Left (a) shows the input image paired with a segmentation map. With semantic conditioning (b), snowfall presents more prominent feature (highlighted in red box) and aligns better with scene structures compared to (c) without it, while cross-attention maps highlight improved view awareness.
    }
    \label{fig:Ablation_seg}
    \vspace{-6mm}
\end{figure}

\noindent \textbf{Semantic Conditioned Input.}
Note that, although the proposed 2D background editing method relies on a semantic map as input (as shown in Figure~\ref{fig:Method_overview}), it only requires the semantic map generated under sunny weather conditions, which can be easily obtained using current open-source models. To evaluate the effectiveness of semantic-conditioned input, we compare editing results and FID scores with and without it. By computing the cross-attention map using ``snowy'' in the text prompt, we observe that semantic conditioning enhances the model’s response to sidewalks and trees, aligning weather effects more accurately with scene structures, as shown in Figure~\ref{fig:Ablation_seg}. Additionally, the model achieves a lower average FID value across different weather conditions compared to cases without semantic conditioning, demonstrating its effectiveness in capturing weather characteristics during training.

\begin{figure}[]
    \centering
    \includegraphics[width=\linewidth]{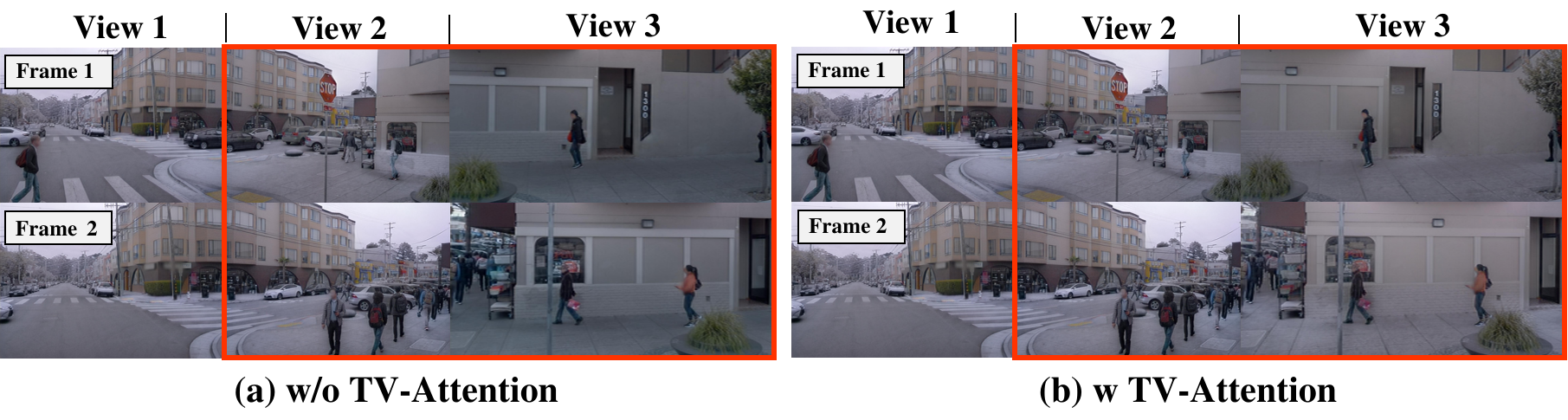}
    \caption{\textbf{Effect of Temporal-View (TV) Attention on Consistency.} Without TV-Attention (a) , inconsistencies appear across frames and viewpoints, as highlighted in the red box. With TV-Attention, weather effects maintain consistent scene appearance across frames and viewpoints.
    }
    \label{fig:Ablation_att}
\end{figure}

\noindent \textbf{Temporal-View Attention.}
As shown in Figure~\ref{fig:Ablation_att}, without TV-Attention, view 3 exhibits inconsistent editing effects, with a darker tone across two frames compared to view 2. After incorporating temporal-view attention, view 3 achieves more consistent editing, displaying a similar snowy effect between adjacent views and frames. For the qualitative analysis shown in Table~\ref{tab:Ablation_att}, temporal attention reduces warp error, indicating improved temporal coherence, while spatial attention results in a lower Bhattacharyya distance, suggesting more consistent editing across viewpoints. By integrating both temporal and view attention, the full implementation achieves the lowest Bhattacharyya distance while maintaining a balanced reduction in warp error, ensuring globally consistent editing results.
\begin{figure}
    \centering
    \includegraphics[width=1\linewidth]{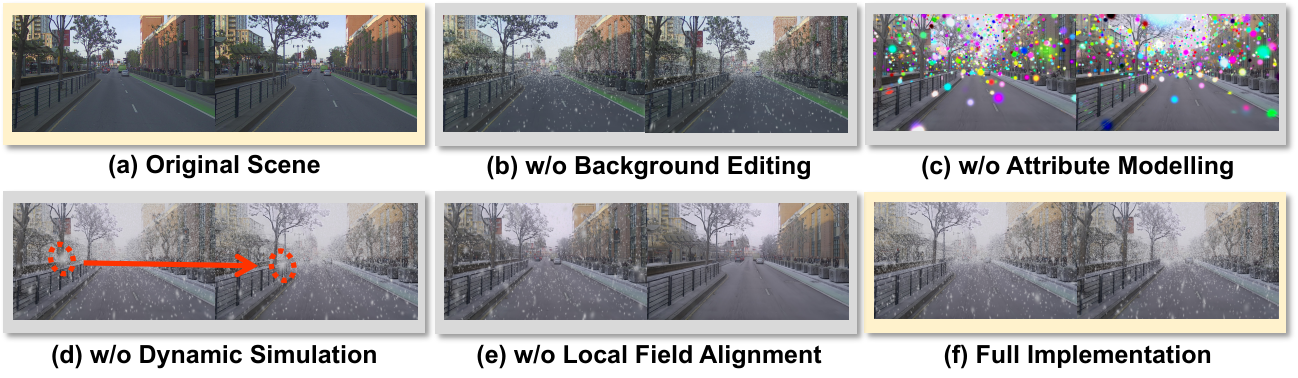}
    \caption{\textbf{Ablation on WeatherEdit modules.}}
    \label{fig:Ablation_3D}
    \vspace{-3mm}
\end{figure} 

\begin{table}[t]
\setlength{\tabcolsep}{4.5pt}  
\centering
\small
\begin{tabular}{lcccc}
\toprule
& \textbf{Self-attn} & \textbf{+View} & \textbf{+Temp} & \textbf{Full} \\
\midrule
Warp-error~($\downarrow$) & 0.043 & 0.042 & \textbf{0.039} & 0.041 \\
Bhattacharyya dist~($\downarrow$) & 0.272 & 0.253 & 0.261 & \textbf{0.245} \\
\bottomrule
\end{tabular}
\caption{\textbf{Ablation on temporal-view attention.}}
\label{tab:Ablation_att}
\vspace{-5mm}
\end{table}

\noindent \textbf{Modular Design for 4D Gaussian Field.}
As the cross-frame results in Figure~\ref{fig:Ablation_3D} show, without 2D background editing (b), the scene lacks an overall weather tone. Randomly initializing 4D Gaussians for attribute modeling (c) produces colorful blobs instead of realistic weather. Without dynamic simulation (d), the particles remain relatively static in the camera view, lacking the natural falling motion. Without local field alignment (e), weather particles stay fixed at the initial position of the scene. As the camera moves forward, the weather particles gradually disappear instead of persisting throughout the scene. In contrast, when all modules are included, the full implementation (f) generates a realistic weather background with natural dynamics.

\section{Conclusion}
In this work, we introduce WeatherEdit, a novel 4D weather editing framework capable of generating realistic weather effects such as fog, rain, and snow, with fine-grained control over their severity. Our approach leverages an all-in-one adapter that enables a diffusion model to transform normal-weather images into diverse weather styles. To maintain consistency of edited results across multiple frames and viewpoints, we introduce an ordered Temporal-View (TV) Attention  mechanism. Additionally, by designing a 4D Gaussian field, we model the attributes and dynamics of weather particles within a 3D scene, enhancing realism and enabling fine-grained control over weather severity. WeatherEdit demonstrates its potential in generating extreme weather conditions from normal scenes, providing a foundation for future research on evaluating the resilience of real-world applications such as autonomous driving.

\bibliography{aaai2026}

\end{document}